\documentclass[conference]{IEEEtran}

\usepackage{array}
\usepackage{amssymb,amsmath}
\usepackage{booktabs}
\usepackage{caption}
\usepackage{color}
\usepackage{graphicx}
\usepackage{framed}
\usepackage{float}
\usepackage{latexsym}
\usepackage{multirow}
\usepackage{times}
\usepackage{url}
\usepackage{threeparttable}
\usepackage{multicol} 

\newcommand\etal{\emph{et~al.}}
\newcommand\ie{\emph{i.e.}}

\def\BibTeX{{\rm B\kern-.05em{\sc i\kern-.025em b}\kern-.08em
    T\kern-.1667em\lower.7ex\hbox{E}\kern-.125emX}}

\begin{document}

\title{Cross-Domain Document Layout Analysis Using Document Style Guide}


\author{\IEEEauthorblockN{Xingjiao Wu$^1$, Luwei Xiao$^2$, Xiangcheng Du$^1$, Yingbin Zheng$^3$, Xin Li$^2$, Tianlong Ma$^1$, Cheng Jin$^1$, Liang He$^2$}
\IEEEauthorblockA{$^1$\textit{Fudan University}~~
$^2$\textit{East China Normal University}~~
$^3$\textit{Videt Lab}}    
}

\UseRawInputEncoding
\maketitle

\begin{abstract}
Document layout analysis (DLA) is a crucial computer vision task that involves partitioning document images into high-level semantic regions such as figures, tables, backgrounds, and texts. 
Deep learning models for DLA typically require a large amount of labeled data, which can be expensive. 
{Though} some researchers use generated data for training, a substantial style gap exists between the generated and target data. 
{Moreover, it is necessary to improve the quality of the generated samples to achieve better control.}
To address these challenges, we propose a cross-domain DLA framework called DL-DSG, which leverages document-style guidance. DL-DSG comprises three components: the document layout generator (DLG) responsible for generating document element locations, the document element decorator (DED) for filling the elements, and the document style discriminator (DSD) for style guidance. In addition to generating controlled documents, we also focus on bridging the gap between the generated and target samples. 
{To this end}, we introduce a novel strategy that transforms document style judgment into the document cross-domain style guidance component.
We evaluate the effectiveness of DL-DSG on popular DLA datasets, including {PubLayNet}, DSSE-200, CS-150, and CDSSE, and demonstrate its superior performance.
\end{abstract}

\begin{keywords}
Data Generation, Document Layout Analysis, Deep Learning, Document cross-domain analysis
\end{keywords}

\section{Introduction}
\label{sec:intro}

Document layout analysis (DLA) is a vital technique for optical character recognition (OCR). DLA has played an important role in table extraction, visual question answering, and information extraction~\cite{binmakhashen2019document}. {More recently, applying deep neural-network-based methods for document layout has become a mainstream {method}~\cite{yan2018fast,wu2020lcsegnet,xu2020layoutlm}.}

\begin{figure}[t]
	\centering
	\includegraphics[width=1\linewidth]{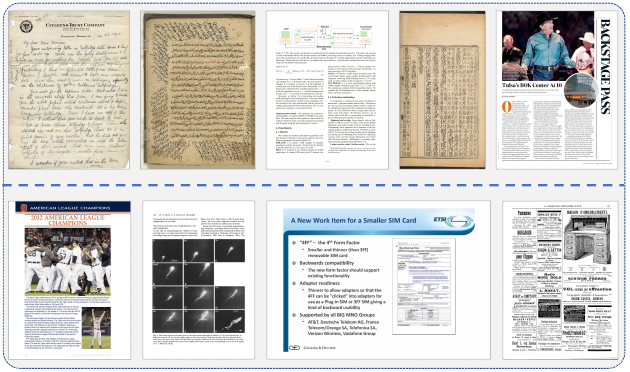}
	\caption{{The first row {contents} comes from different data sets and {those of the second row are from} the same data set} (DSSE-200~\cite{yang2017learning}).}
\label{F_motivation}
\end{figure}

However, {in real-world scenarios}, the document layout analysis needs to deal with various styles (\ie,{historical handwritten documents}~\cite{xu2018multi}, {Arabic historical documents}~\cite{bukhari2012layout}, {academic documents}~\cite{clark2015looking}, {Asian languages documents}~\cite{shen2020large}, {magazine documents}~\cite{clausner2017icdar2017}, etc. (Fig.~\ref{F_motivation} first rows)).
{A {huge gap} exists between individual documents, even from the unity style documents (Fig.~\ref{F_motivation} second rows).} The DLA is a pixel-level document classification task, and differences in a single object often significantly impact the model outcomes.

{Numerous researchers are currently investing considerable efforts in leveraging large-scale pre-training models to tackle the challenge of unifying document styles. 
They achieve this by enhancing data through generative approaches.
{Several notable {studies} have recently emerged, particularly those based on {generative adversarial networks} (GANs). 
However, these frameworks primarily focus on generating a series of layout boxes without utilizing them to compose complete documents with all elements. {In addition, previous studies} have emphasized generating more style layout boxes {; however, they} have not adequately addressed evaluating the quality of these layout boxes.
Furthermore, ensuring that the generated documents {fulfill} high-quality standards to enhance the model's generalization capabilities is crucial.} Without fine-grained control over the quality of generated documents, the model may learn too many invalid layouts, which can ultimately hamper its performance.}

{{{To} enhance the robustness and generalization {capability of document layout models, two pivotal facets must be considered.} 
{First}, guaranteeing the quality of the training samples is of utmost importance. 
Secondly, it is imperative to establish a connection between {the training and target samples. Numerous researchers have proposed pre-training methodologies to address this disparity.} 
Nevertheless, these approaches frequently necessitate fine-tuning with supplementary annotation data. 
{However, manual annotation is expensive} and inconvenient for model automation.}
Therefore, two key questions arise: 
(1) How can we incorporate an unsupervised quality assessment component into the document layout generation process? (2) What is an effective strategy {for reducing} the gap between training and target samples in {document layout analysis} (DLA)?}
Specifically, {the first question can be considered as how} to create a method to reduce the gap between the generated sample and the high-quality document layout sample.
{Overall,} the core question of this {study} is \emph{ How do we enhance the generalization of the model by creating a method with less manual intervention and solve the cross-domain problem in document layout analysis?}

{{Data generation has emerged as an effective solution to address cross-domain DLA challenges.} {However, {a significant limitation} of these methods {is} the stylistic differences between the generated and target samples.} This challenge hinders the performance of models trained on generated data when applied to target samples.
{To {overcome} this limitation, it is {necessary to develop} a technique that better steers the generated exemplars to emulate {the stylistic attributes of the target data.}} {Minimizing the} need for excessive manual intervention in this method is essential to enhance its practicality and efficiency.
In this context, contrastive learning has shown promising potential {owing to} its superior generalization ability, as it learns representations by comparing positive and negative pairs~\cite{nan2021interventional}. 
{Further,} contrastive learning focuses on distinguishing data at the abstract semantic level in the feature space, thereby bypassing the need for tedious details of individual {examples to overcome} this {imitation. It} is crucial to develop a method that guides the generated samples to align styles more closely with the target samples.}

{We introduce the document layout-generator decorator discriminator (DL-GDD) framework to achieve practical cross-domain document layout analysis with style guidance. The DL-GDD comprises three essential components: document layout generator (DLG), document element decorator (DED), and document style discriminator (DSD). DLG is responsible for generating document layout boxes, whereas DED uses these layout boxes to create complete document pages following aesthetic norms. DSD leverages a fusion of comparison learning methods to guide the style of generated data. Our contributions can be summarized as follows.}

\begin{itemize}
	\item {We propose an unsupervised framework that integrates {cross-domain document layout analysis and document layout quality assessment tasks.}}
	\item {Our methodology introduces an unsupervised strategy to mitigate cross-domain disparities in document layout analysis, eliminating the need for annotated data. Furthermore, we extend the applicability of our method to document layout quality assessment, thereby establishing it as a pioneering, unsupervised approach for this task.}
	\item  {We devised a document element decorator to generate document pages using layout boxes.} This method realizes document elements decorated to follow the specifications of aesthetic methods. {Moreover, the document element decorator supports document quality assessment.}
	\item Experimental results on four widely-used public datasets verify that our method is efficacious and robust for document quality assessment and cross-domain document layout analysis. 
\end{itemize}

\section{Related Work}
\label{sec:related}

\vspace{0.08in}
\noindent\textbf{{Document Layout Analysis.}}
{Processing methodologies for document layout analysis (DLA) can be classified into two categories: traditional and deep learning-based. Traditionally, the former utilizes image segmentation algorithms to partition the layout. However, as the complexity of page layouts has escalated, traditional DLA methods have become inadequate for satisfying the processing requirements. Consequently, the focus has shifted toward deep-learning-based approaches, which have gained considerable attention. Presently, deep learning-based methods not only emphasize the division of high-level semantic areas within documents, but also emphasize the model’s processing speed and generalization capabilities.}
Lu~\etal~\cite{lu2021probabilistic} {utilized the unique perception and recognition abilities of humans in text regions in complex layouts and introduced a document segmentation probability framework. It is worth mentioning that they conceptualized text homogeneity into gestalts displayed in text regions. In recent years, pre-training technology has been widely used and has had an irreplaceable effect.}
{To} make full use of the document information, Xu~\etal~\cite{xu2020layoutlm} {proposed LayoutLM to jointly model the relationship between the text in a document and the layout information.}
Studer~\etal~\cite{studer2019comprehensive} {found that ImageNet's pretrained DLA model can help analyze historical documents.}
{To consistently extract high-quality text from formatted PDFs,} Soto~\etal~\cite{soto2019visual} devised an object detection technology with enhanced context features to segment documents.
{The study conducted by} Minouei~\etal~\cite{minouei2021document} {was no longer limited to the application of FCN-based semantic segmentation methods. They used a deep neural network inspired by natural-scene object detectors for document layout analysis.}
Xu~\etal~\cite{xu2018multi} {focused on the relevance of document layout analysis tasks. They trained a multitask FCN for document layout analysis.}
Ma~\etal~\cite{ma2020joint} trained an end-to-end trainable framework to process historical document content. They combined character detection and recognition with layout analysis {in their framework}.

Recent {studies have considered the} man-machine hybrid method for document layout analysis~\cite{wu2021human}. 
{Zaragoza}~\etal ~\cite{calvo2017music} {used a manual-assisted calculation method and analyzed the layout of music files.}
{Oliveira~\etal~\cite{augusto2017fast} used a Fast CNN for document-layout analysis to improve the efficiency of the DLA model.}
{Patil~\etal~\cite{patil2021layoutgmn} leveraged a graph matching network (GMN) to predict the structural similarities between two-dimensional layouts. Yang~\etal~\cite{yang2017learning} proposed an end-to-end, multimodal, fully convolutional network to extract semantic structures from document images.}

{These methodologies play a pivotal role in document layout analysis, exerting profound influences on the advancement and progression of document layout analysis tasks.}

\vspace{0.08in}
\noindent\textbf{Document Layout Generation and Document Quality Assessment.}
{With the development of office automation, document styles are emerging increasingly, and their designs are becoming increasingly complex. The current amount of labeled data is far less than the amount of data generated. Therefore, several researchers have started exploring the use of data synthesis for document layout analysis. }
Yang~\etal~\cite{yang2017learning} analyzed data with the help of Latex synthetic document layout{.} 
Li~\etal~\cite{li2021harmonious} used deep aesthetics to guide and generate a more realistic text layout{.} 
Tabata~\etal~\cite{tabata2019automatic} explored the automatic layout generation of graphic design magazines{.} 
Dayama~\etal~\cite{dayama2020grids} designed an interactive layout mode with integer programming{.} 
Lee~\etal~\cite{lee2020neural} also explored the graphic layout generation with constraints{.} 
Zhong~\etal~\cite{zhong2019publaynet} generated data by automatically analyzing the {PDF} content publicly available on the PubMed center{.} Li~\etal~\cite{li2019layoutgan,li2020layoutgan} exploited {LayoutGAN} for sample generation{.} 
Arroyo~\etal~\cite{arroyo2021variational} applied a variational transformer network for layout generation. 
Kikuchi~\etal~\cite{kikuchi2021constrained} generated a graph layout {using} potential optimization constraints.

\vspace{0.08in}
\noindent\textbf{{Cross-Domain Document Layout Analysis.}}
{Because the data used for pretraining are essentially different from the target data, cross-domain technologies must be considered. In recent years, several researchers have focused intensively on cross-domain research in document layout analysis.}
Li~\etal~\cite{li2020cross} devised a benchmark test suite and method for cross-domain document object detection, and Singh~\etal~\cite{singh2020multi} introduced a method for understanding {multidomain documents using the} few-shot object {detection layout method.} 
Goslin~\etal~\cite{goslin2013cross} put forward a cross-domain evaluation method for document-to-HTML conversion tools to quantify the loss of text and structure {during} document analysis.
The current document quality {evaluation study focuses} on evaluating document content and image quality~\cite{alaei2018blind,rodin2021document,rai2018document,li2017attention,lu2019deep,kang2014deep}, {and there is a lack of studies that specifically evaluate the quality of document layout.}

{The aforementioned methods have had a significant impact on the advancement of document layout generation research. Nevertheless, certain GAN-based approaches primarily focus on generating document layout boxes while neglecting the task of populating document elements. This limitation hinders the evaluation of the quality of the generated document pages. Our study aimed to address this deficiency and fill this research gap.}

\section{Approach}
\label{sec:method}
\begin{figure*}[t]
	\centering
		\includegraphics[width=1\linewidth]{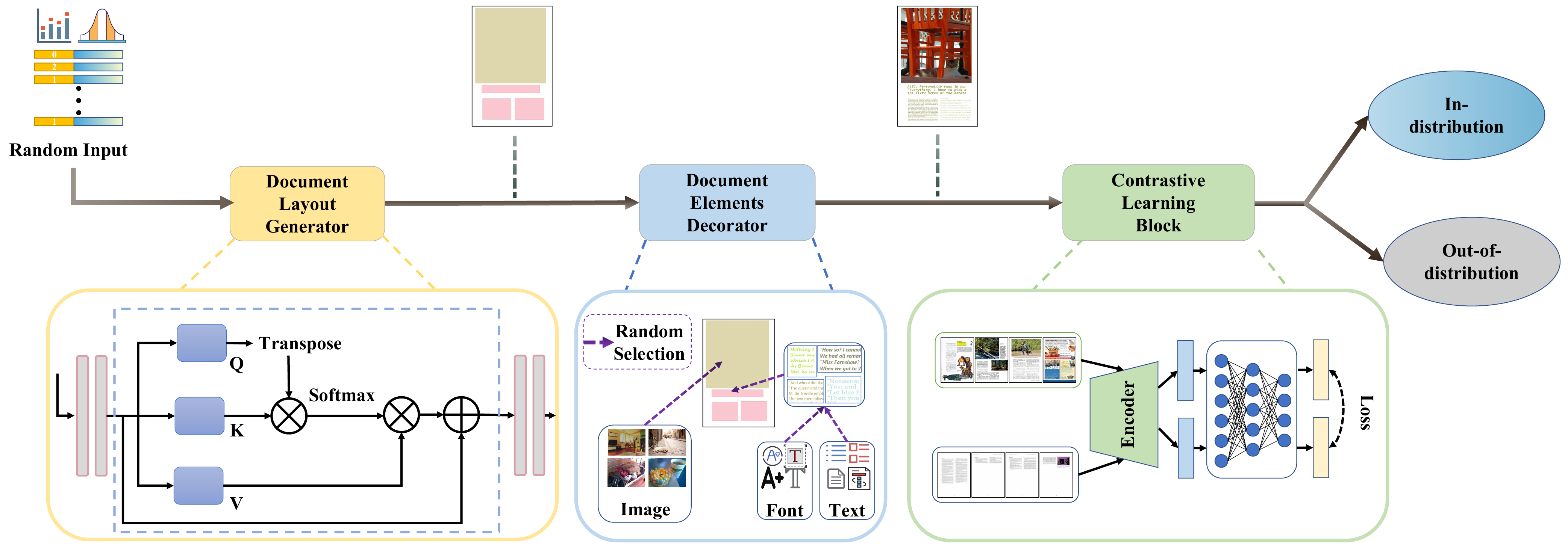}
		\caption{Illustration of the DL-GDD.}
		\label{F_framework}
	\end{figure*}

{The main objective of DL-GDD is to generate samples that closely resemble the target samples. 
To achieve this goal, we first employ a LayoutGAN-based~\cite{li2019layoutgan,li2020layoutgan} approach to {construct document layout generator} (DLG), which is responsible for generating document layout boxes. 
{Next, it is essential to produce complete documents to {ensure the alignment} between the generated and target samples. }
To accomplish this, we introduce the {document element decorator}(DED). The DED utilizes aesthetic norms and specifications to generate complete document {pages based on the} document layout boxes.
Finally, we address the incorporation of style guidance for the generated data through the {design of a document style discriminator}(DSD). By integrating contrastive learning methods{, DSD ensures that the generated data effectively align with the desired styles.}}

\subsection{Network Architecture}

{{An overview of the DL-GDD framework is shown in} Fig.~\ref{F_framework}, {comprising three essential components DLG, DED, and DSD. First, the DLG is trained using document datasets available on the Internet, such as PubLayNet and Magazine. DED is rule-driven and does not require separate training. After training the DLG, the network parameters are frozen, and the DSD is trained. The training labels for DSD do not require fine-grained annotation; simple data differentiation is sufficient, such as distinguishing between high-quality or low-quality layouts or approximate categorization like DSSE-200, CS-150, or CDSSE. Upon completing the training, inputting random vectors into the system produces document images accompanied by quality classification labels.}}

{The entire process can be formalized as follows{.} The input of DL-GDD is a set of randomly coded vectors, where each vector represents a layout on a document page. Each vector consists of five dimensions: the class of the layout box ($c$), the $x$ coordinate of the upper-left corner of the layout box, the $y$ coordinate of the upper-left corner of the layout box, the width of the layout box ($w$), and the height of the layout box ($h$).
{If} V {represents} the input random variable, then ${\mathop{\rm V}\nolimits}  = \{ {v_i}\} _{i = 1}^N$, where ${v_i} = \{ {c_i},{x_i},{y_i},{w_i},{h_i}\}$, and $\{ {x_i},{y_i},{w_i},{h_i}\}  \sim {\mathcal N}(0,I)$, which denotes that the values of ${x_i},{y_i},{w_i}$, and ${h_i}$ are drawn from a normal distribution {with a mean of zero and identity} covariance matrix $I$.}

The input vector V is transformed into a new page layout box $V'$ using a trained DLG. The DLG is defined as: $G:V \mapsto V'$.
Next, we employ a document {element} decorator to fill in the document elements; the DED is defined as: $D:V' \mapsto P$.
Finally, we incorporate a contrastive learning block for quality assessment. We define the DSD as: $SD:P \mapsto {P_q}$.
Moreover, our document style discriminator can serve as document cross-domain style guidance. 
To achieve this, we only need to replace the data {in} the contrastive learning block. 
The document cross-domain style guidance is defined as: $O:{P_q} \mapsto {P_s}, {P_s} \in {P_q}$.
{Consequently, our model produces a series of relevant documentation pages as outputs.}

\subsection{Document Layout Generator}

{A document layout generator (DLG) is a generative adversarial network comprising a generator and discriminator with similar architectures.
First, the generator uses two fully connected units to encode the data. The encoded data are then passed through a transform block~\cite{vaswani2017attention} to obtain a feature vector that captures the relationships between the elements. Next, the feature vector is decoded using two fully connected units to derive the position code for the current page.
Subsequently, the discriminator identifies the generated position code.
As defined in Section 3.1, the DLG is defined as: $G:V \mapsto V'$, where $V$ represents the input vector and $V'$ represents the restored new page layout box.
The DLG process is represented by Eq.~\ref{equ:Gen}.

\begin{equation}
\begin{array}{c}
	V \mapsto V' \Leftrightarrow \left\{ {\begin{array}{*{20}{l}}
		{{\rm{V}} = \{ {v_i}\} _{i = 1}^N}\\
		{{\rm{V'}} = \{ {v_i}^\prime \} _{i = 1}^N}\\
		{{v_i}^\prime {\rm{ = }}g(\varphi (f({v_i};{\theta _1});{\theta _2});{\theta _3})}
		\end{array}} \right.
\end{array}
\label{equ:Gen}
\end{equation}
where $f$ and $g$ are multilayer perceptrons. Here, $f$ representing the encoding operation of the input information, and $g$ represents the decoding operation represents the intermediate features. $\theta_1, \theta_2, \theta_3,$ denotes the training parameter set of the generator, whereas $\varphi$ involves the use of the transformer block.
Through a series of continuous operations, we reconstruct the original input vector $v$ into a new set of page elements $v'$.
Finally, we restore the element collection to form a document page layout.

The DLG calculates the loss based on the similarity between two sets of bounding boxes and labels. We use the mean squared error to measure the loss on the bounding boxes and cross-entropy to measure the loss on the labels.

\begin{equation}
	\begin{array}{l}
		{\cal L} = \frac{1}{{\rm{N}}}{\sum {\left( {{v_i}^\prime  - {v_i}} \right)} ^2} - \sum {{y_i}\log ({p_i})}
	\end{array}
	\label{equ:DLGLoss}
	\end{equation}
where $y_i$ represents the ground truth labels, and $p_i$ is the predicted label.

\subsection{Document Elements Decorator}
{The document element decorator (DED) involves the process of filling elements, such as figures, tables, and text, into the document layout to create the final document page. Aesthetics guide this process to ensure a visually pleasing composition. Document layout analysis (DLA) aims to identify high-level semantic areas in document pages. These semantic areas include figures, tables, and text. From a compositional perspective, figures and tables can be grouped as figure areas. However, text is considered separately, with factors such as text size and color source. As shown in} Fig~\ref{F_framework}, we divide the content into figure and text components. For the figure component, we use the COCO dataset~\cite{lin2014microsoft} as the source. {We randomly selected images from the dataset, but because it is challenging to identify images with the exact dimensions required for pasting, we needed to crop the images. To enhance the visual appeal of the images, certain criteria were set for their dimensions. Specifically, the image width should be greater than the width of the pasted area and less than 1.5 times the width of the pasted area. Further, the image height should be greater than the height of the pasted area and less than twice that of the pasted area. We used a series of novel English texts as the source of the text component. We then randomly selected fonts and colors for the combination. The text font size is selected from the range} $[8, L_h]$, where $L_h$ is half the height of the pasted area. {This ensured that the text was appropriately sized in the figure area. As defined in Section} 3.1, the DED is defined as:  $D:V' \mapsto P$.
{The DED process is represented by Eq.~\ref{equ:Ded}.}

\begin{tiny}
\begin{equation*}
\begin{array}{l}
	V' \mapsto P \Leftrightarrow \left\{ {\begin{array}{*{20}{l}}
		{{\rm{V'}} = \{ {v_i}^\prime \} _{i = 1}^N}\\
		{P = \{ {p_i}\} _{i = 1}^N}\\
		{{p_i}{\rm{ = }}{v_i} + \{ i{m_{i0}}, \cdots ,i{m_{ia}}\}  + \{ tei{m_{i0}}, \cdots ,tei{m_{ib}}\} }\\
		{\left\{ {\begin{array}{*{20}{l}}
		{a > 0,b > 0}\\
		{{w_{i*}} \le {\Gamma _w}\left( {i{m_{i*}}} \right) \le 1.5 \times {w_{i*}}}\\
		{{h_{i*}} \le {\Gamma _h}\left( {i{m_{i*}}} \right) \le 2 \times {h_{i*}}}\\
		{tei{m_{ib}} = \left\{ {\xi \left( {te} \right) + \xi \left( {font} \right) + \xi \left( {color} \right)} \right\}}\\
		{{\Gamma _w}\left( {tei{m_{i*}}} \right) \ge {w_{i*}}}\\
		{{\Gamma _h}\left( {tei{m_{i*}}} \right) \ge {h_{i*}}}
		\end{array}} \right.}
		\end{array}} \right.
\end{array}
\label{equ:Ded}
\end{equation*}
\end{tiny}

{where $i{m_{ia}}$ indicates the $a-th$ box candidate image of the $i-th$ document page;
$tei{m_{ib}}$ represents the $b-th$ box candidate text image of the $i-th$ document page;
$\Gamma _w$ refers to the width of the calculation object;
$\Gamma _h$ is the height of the calculation object;
$\xi$ is the random selection of an element from a set,
$te$ means a collection of candidate pictures (COCO data set);
$font$ represents a candidate font library;
$color$ stands for a combination of candidate colors;
$w_{i*}$ represents the width of the current paste area, and $h_{i*}$ represents the height of the current paste area.}

\subsection{Document Style Discriminator}

{The document style discriminator (DSD) is an unsupervised component based on contrastive learning. We implemented DSD to identify document styles. In this study, the DSD identifies two styles. Our contrast-learning method was inspired by SupContrast}~\cite{khosla2020supervised}.
{We define unlabeled target domain document images as positive samples, and define Workshop papers with visually discernible differences under probability statistics as negative samples, the Workshop papers from CVPR (Conference on Computer Vision and Pattern Recognition), ICCV (International Conference on Computer Vision), ECCV (European Conference on Computer Vision), and ACM MM (ACM International Conference on Multimedia)  from 2018 to 2020~\cite{von2010paper}.}
{As defined in section 3.1, the DSD is defined as:  $D:P \mapsto P_q$.
{The DSD process is represented by} Eq.~\ref{equ:Ded2}.}

\begin{tiny}
\begin{equation*}
	P \mapsto {P_q} \Leftrightarrow \left\{ {\begin{array}{*{20}{l}}
		{P = \{ {p_i}\} _{i = 1}^N}\\
		{{P_q} = \{ p_{_q}^ + \} _{i = 1}^{ \le N} + \{ p_{_q}^ - \} _{i = 1}^{ \le N}}\\
		{{p_q} = \left\{ {\begin{array}{*{20}{c}}
		{p_{_q}^ + }&{}&{score(\gamma ({p_i}),\gamma ({S^ + })) < score(\gamma ({p_i}),\gamma ({S^ - }))}\\
		{p_{_q}^ - }&{}&{score(\gamma ({p_i}),\gamma ({S^ + })) > score(\gamma ({p_i}),\gamma ({S^ - }))}
		\end{array}} \right.}
		\end{array}} \right.
\label{equ:Ded2}
\end{equation*}
\end{tiny}

{where $\gamma$ represents feature extraction, $p_i$ is the currently generated document page, $S^ +$ indicates a set of positive samples, $S^ -$ refers to a set of negative samples, and $score$ stands for a similarity score.}

The DSD utilizes the loss function shown in Eq.~\ref{equ:DedLoss} for constraint, where the data examples are obtained from the document examples generated by DED and denoted as $P$.
 $z$ represents the feature vector after encoding projection, and $\cdot$ denotes the inner (dot) product. 
$\tau $ is the scalar temperature parameter, $j(.)$ represents a positive sample, and $A(i) \equiv {P \mathord{\left/ {\vphantom {P {\{ p\} }}} \right. \kern-\nulldelimiterspace} {\{ p\} }}$.

 \begin{equation}
	\mathcal L = \sum\limits_{p \in P} {{\mathcal L _p}}  =  - \sum\limits_{p \in P} {\log \frac{{\exp ({z_p}\cdot{{{z_{j(p)}}} \mathord{\left/
 {\vphantom {{{z_{j(p)}}} \tau }} \right.
 \kern-\nulldelimiterspace} \tau })}}{{\sum\limits_{a \in A(p)} {\exp ({z_p}\cdot{{{z_a}} \mathord{\left/
 {\vphantom {{{z_a}} \tau }} \right.
 \kern-\nulldelimiterspace} \tau })} }}}
\label{equ:DedLoss}
\end{equation}

\begin{figure}[t]
	\centering
	\includegraphics[width=\linewidth]{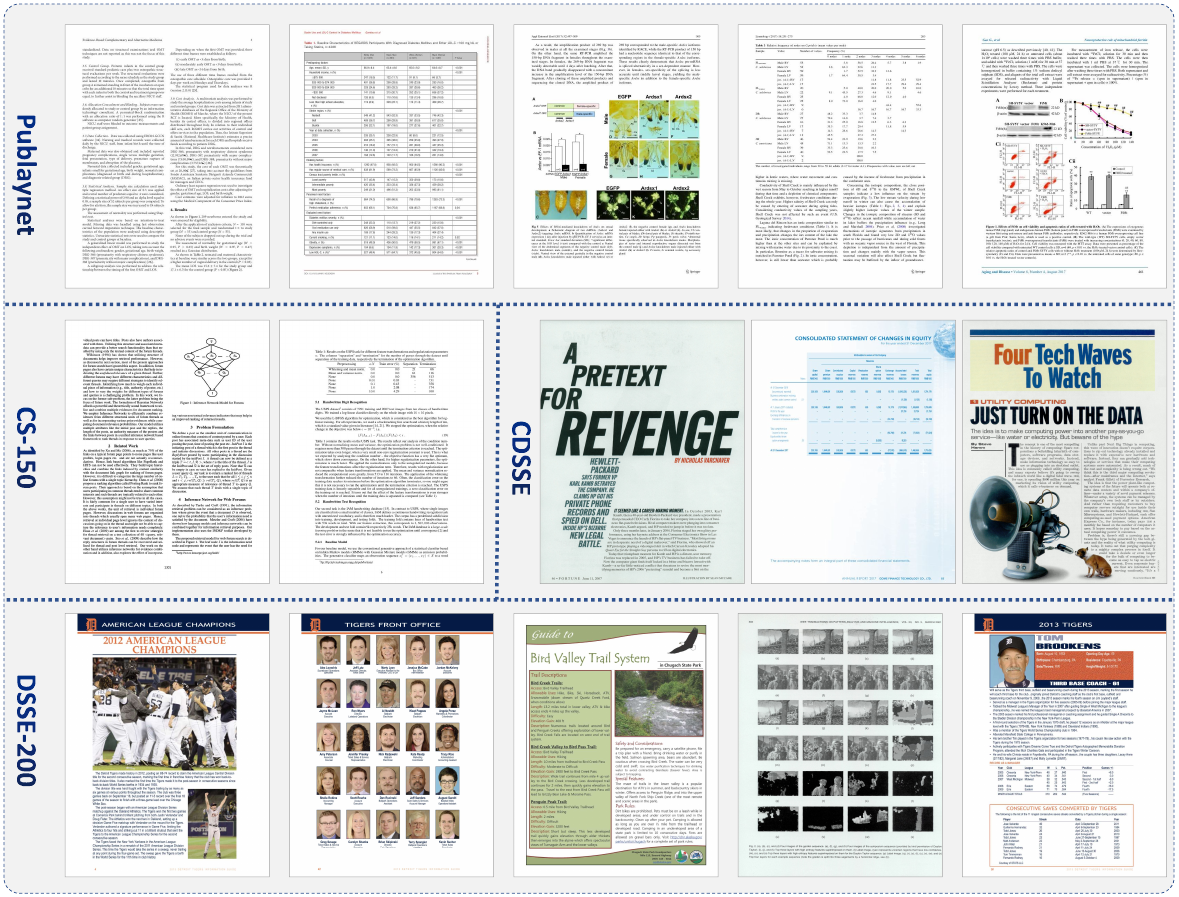}
	\caption{An example of the dataset.}
	\label{F_Dataset}
\end{figure}

\vspace{0.08in}
\noindent
\textbf{The style discriminator for document quality assessment.}
Document layout data generated from a random Gaussian distribution can exhibit considerable variability in quality, thereby introducing noise during the model training phase and adversely affecting the model performance. Consequently, there is an urgent need for a universal approach for assessing the quality of document layouts. Contrastive learning, which involves acquiring representations by comparing positive and negative pairs, has emerged as a promising solution. By applying contrastive learning, it is feasible to distinguish between high- and low-quality document layouts by leveraging their respective representations. However, determining methods to effectively measure and distinguish the quality of document layouts remains a key challenge. This assessment is crucial for guiding the model towards generating high-quality documents and improving its overall performance. Generally, magazine layouts are designed and typeset, and the quality of the layout is higher than that of conference workshop papers.
Therefore, we collected journal pages from Reader's Digest and China's National Geographic as positive samples and collected the workshop papers as negative samples.

\vspace{0.08in}
\noindent
\textbf{The style discriminator for cross-domain.}
The development of a universal document layout analysis (DLA) model is a prominent area of research for many scholars. 
However, the rapid pace of document generation often surpasses the update speeds of the models and datasets. This has led researchers to explore cross-domain document layout analysis as a practical approach to address this challenge. The cross-domain DLA enhances the generalization of the model by incorporating more relevant data from different domains. Despite these efforts, a considerable gap often exists between the training and target data, which can hinder the effectiveness of a model optimized using the training data. Ideally, the model should directly leverage the target data for training to achieve optimal results. However, if the model continuously invests in the target data, its generalization weakens and labor costs increase. We generated sample data from a Gaussian distribution; therefore, if we can find certain distributions similar to the target dataset distribution and expand them, we can create a more suitable sample. If we obtain these data, we put them into the DL-GDD for iteration. We generated sufficient samples close to the target dataset and built a bridge between the training and target data. We used data from the unlabeled target datasets (\ie, DSSE-200, CS-150, and CDSSE) as positive samples and data from the workshops as negative samples.

\section{Experiments}
\label{sec:exp}
In this section, we introduce the implementation details of the DL-GDD framework, which covers parameter selection and training specifications. Subsequently, we present the datasets used for the experimental validation and compare the differences between them, highlighting the necessity of conducting a cross-domain document layout analysis. Following that, we analyze the results, with a specific focus on the performance of DL-GDD in handling academic documents (CS-150), comprehensive documents (DSSE-200), and complex non-Manhattan layout documents (CDSSE). We provide intuitive visualizations of the results for comparative analysis. Finally, ablation studies are conducted for each DL-GDD component to demonstrate the effectiveness of the proposed components.

\begin{table*}[t]
	\centering
	\caption{Cross-domain document layout analysis results between {PubLayNet} to CS-150.  The ``DA'' means use all synthetic data; The ``DQA'' means use document quality assessment; The ``CD'' means use the style discriminator for cross-domain. The best results {are shown as boldface text.} }
	\label{T_CS}
	\scalebox{0.92}{
	\begin{tabular}{p{120pt}p{16pt}<{\centering}p{16pt}<{\centering}p{16pt}<{\centering}p{16pt}<{\centering}||p{16pt}<{\centering}p{16pt}<{\centering}p{16pt}<{\centering}p{16pt}<{\centering}||p{16pt}<{\centering}p{16pt}<{\centering}p{16pt}<{\centering}p{16pt}<{\centering}}
	\toprule
	\multirow{2}{*}{} & \multicolumn{4}{c||}{{PubLayNet} $ \to $ CS-150 \& DA} & \multicolumn{4}{c||}{{PubLayNet} $ \to $ CS-150 \& DQA} & \multicolumn{4}{c}{{PubLayNet}  $ \to $ CS-150 \& CD} \\ \cline{2-13}
	
			 & A         & P         & R        & F1        & A                & P                & R                & F1                & A             & P             & R            & F1            \\    \midrule	
	FCN/VGG & 0.956 & \textbf{0.883} & 0.597 & 0.712 & 0.988 & 0.882 & 0.731 & 0.799 & 0.978 & \textbf{0.886} & \textbf{0.821} & \textbf{0.852} \\
	FCN/R18~\cite{wu2021document} & 0.752 & 0.659 & 0.525 & 0.584 & 0.941 & 0.838 & 0.592 & 0.694 & 0.966 & 0.879 & 0.706 & 0.783 \\
	FCN/R50 & 0.936 & 0.826 & 0.649 & 0.727 & 0.969 & 0.854 & 0.725 & 0.784 & 0.982 & 0.865 & 0.759 & 0.809 \\
	PSPnet~\cite{zhao2017pyramid}   & 0.889 & 0.752 & \textbf{0.756} & 0.754 & 0.965 & 0.810 & 0.725 & 0.765 & 0.973 & 0.852 & 0.700 & 0.769 \\
	PANet~\cite{li2018pyramid}   & 0.921 & 0.789 & 0.612 & 0.689 & 0.939 & 0.881 & 0.601 & 0.715 & \textbf{0.988} & 0.812 & 0.660 & 0.728 \\
	DV3+~\cite{chen2018encoder}    & 0.802 & 0.701 & 0.512 & 0.592 & 0.815 & 0.732 & 0.511 & 0.601 & 0.922 & 0.802 & 0.533 & 0.640 \\
	DV3+/R18~\cite{chen2018encoder} & 0.946 & 0.799 & 0.693 & 0.742 & 0.979 & 0.881 & 0.799 & 0.838 & 0.985 & 0.879 & 0.742 & 0.805 \\
	DRFN~\cite{wu2021document}     & \textbf{0.966} & 0.825 & 0.733 & \textbf{0.776} & \textbf{0.989} & \textbf{0.885} & \textbf{0.802} & \textbf{0.841} & \textbf{0.988} & 0.884 & 0.796 & 0.837 \\ \bottomrule
	\end{tabular}
	}
	\end{table*}

	\begin{figure}[t]
		\centering
		\includegraphics[width=\linewidth]{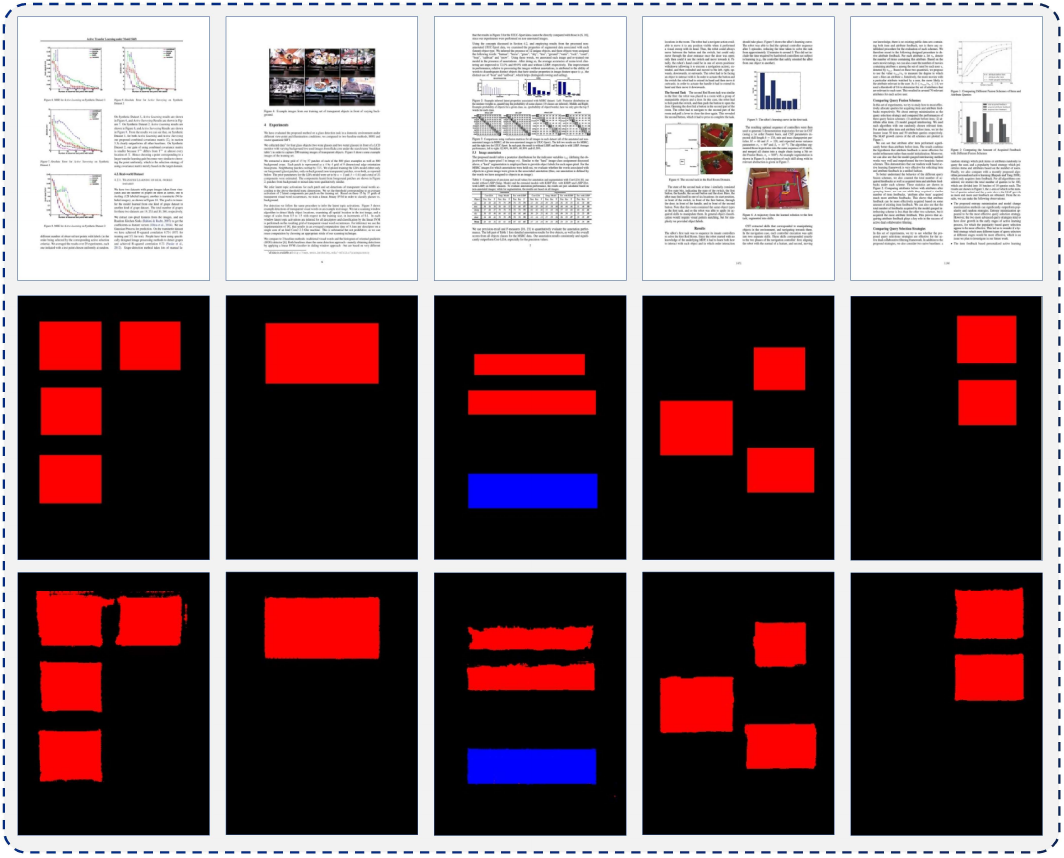}
		\caption{ The CS-150 real documents and their corresponding segmentation. Top: original. Middle: ground-truth. Bottom: predictions. Segmentation label colors are: \colorbox{red}{figure}, \colorbox{blue}{\color{white}{table}}, and \colorbox{black}{\color{white}{No-Image}}.
		}
		\label{F_RCS}
	\end{figure}
	
	\begin{table*}[t]
	\centering
	\caption{Cross-domain document layout analysis results between {PubLayNet and} DSSE-200. The best results are in bold. The ``DA'' means use all synthetic data; The ``DQA'' means use document quality assessment; The ``CD'' means use the style discriminator for cross-domain. The best results {are shown as boldface text.} }
	\label{T_DSSE}
	\scalebox{0.92}{
	\begin{tabular}{p{120pt}p{16pt}<{\centering}p{16pt}<{\centering}p{16pt}<{\centering}p{16pt}<{\centering}||p{16pt}<{\centering}p{16pt}<{\centering}p{16pt}<{\centering}p{16pt}<{\centering}||p{16pt}<{\centering}p{16pt}<{\centering}p{16pt}<{\centering}p{16pt}<{\centering}}
	\toprule
	\multirow{2}{*}{} & \multicolumn{4}{c||}{\footnotesize{{PubLayNet} $ \to $ DSSE-200 \& DA}} & \multicolumn{4}{c||}{\footnotesize{{PubLayNet} $ \to $ DSSE-200 \& DQA}} & \multicolumn{4}{c}{\footnotesize{{PubLayNet} $ \to $ DSSE-200 \& CD}} \\ \cline{2-13}
	
			 & A         & P         & R        & F1        & A                & P                & R                & F1                & A             & P             & R            & F1            \\    \midrule	
	FCN/VGG & 0.751 & \textbf{0.692} & \textbf{0.682} & \textbf{0.687} & 0.813 & \textbf{0.724} & 0.753 & \textbf{0.738} & 0.832 & 0.768 & 0.766 & 0.767
	\\
	FCN/R18~\cite{wu2021document} & 0.423 & 0.435 & 0.554 & 0.487 & 0.742 & 0.660 & 0.601 & 0.629 & 0.812 & 0.793 & 0.712 & 0.750 \\
	FCN/R50 & 0.695 & 0.533 & 0.528 & 0.530 & 0.805 & 0.681 & 0.677 & 0.679 & 0.864 & 0.760 & 0.705 & 0.731 \\
	PSPnet~\cite{zhao2017pyramid}   & 0.685 & 0.408 & 0.496 & 0.448 & 0.787 & 0.680 & 0.680 & 0.680 & 0.811 & 0.724 & 0.733 & 0.728 \\
	PANet~\cite{li2018pyramid}   & 0.664 & 0.558 & 0.563 & 0.560 & 0.733 & 0.535 & 0.649 & 0.587 & 0.829 & 0.695 & 0.659 & 0.677 \\
	DV3+~\cite{chen2018encoder}    & 0.547 & 0.422 & 0.470 & 0.445 & 0.579 & 0.510 & 0.607 & 0.554 & 0.656 & 0.561 & 0.597 & 0.578 \\
	DV3+/R18~\cite{chen2018encoder} & 0.703 & 0.582 & 0.622 & 0.601 & 0.809 & 0.608 & 0.730 & 0.663 & 0.851 & 0.677 & 0.767 & 0.719 \\
	DRFN~\cite{wu2021document}    & \textbf{0.789} & 0.673 & 0.652 & 0.662 & \textbf{0.821} & 0.672 & \textbf{0.785} & 0.724 & \textbf{0.868} & \textbf{0.796} & \textbf{0.801} & \textbf{0.798} \\ \bottomrule
	\end{tabular}
	}
	\end{table*}

	\begin{table*}[t]
	\centering
	\caption{Cross-domain document layout analysis results between {PubLayNet and} CDSSE. The best results are in bold. The ``DA'' means use all synthetic data; The ``DQA'' means use document quality assessment; The ``CD'' means use style discriminator for cross-domain. {The best results are shown as boldface text.} }
	\label{T_CDSSE}
    \scalebox{0.92}{
		\begin{tabular}{p{120pt}p{16pt}<{\centering}p{16pt}<{\centering}p{16pt}<{\centering}p{16pt}<{\centering}||p{16pt}<{\centering}p{16pt}<{\centering}p{16pt}<{\centering}p{16pt}<{\centering}||p{16pt}<{\centering}p{16pt}<{\centering}p{16pt}<{\centering}p{16pt}<{\centering}}

	\toprule
	\multirow{2}{*}{} & \multicolumn{4}{c||}{{PubLayNet} $ \to $ CDSSE \& DA} & \multicolumn{4}{c||}{{PubLayNet} $ \to $ CDSSE \& DQA} & \multicolumn{4}{c}{{PubLayNet} $ \to $ CDSSE \& CD} \\ \cline{2-13}
	
			 & A         & P         & R        & F1        & A                & P                & R                & F1                & A             & P             & R            & F1            \\    \midrule	
	FCN/VGG & 0.632 & 0.657 & 0.652 & 0.654 & 0.688 & 0.716 & 0.768 & 0.741 & \textbf{0.721} & 0.733 & 0.786 & 0.759 \\
	FCN/R18~\cite{wu2021document} & 0.512 & 0.477 & 0.543 & 0.507 & 0.645 & 0.682 & 0.597 & 0.637 & 0.701 & 0.731 & 0.765 & 0.747 \\
	FCN/R50 & 0.624 & 0.618 & 0.627 & 0.622 & 0.678 & 0.701 & 0.749 & 0.724 & 0.699 & 0.725 & 0.761 & 0.743 \\
	PSPnet~\cite{zhao2017pyramid}   & 0.628 & 0.519 & 0.462 & 0.489 & 0.680 & 0.695 & 0.716 & 0.705 & 0.698 & 0.711 & 0.731 & 0.721 \\
	PANet~\cite{li2018pyramid}   & 0.630 & 0.501 & 0.526 & 0.513 & 0.651 & 0.690 & 0.584 & 0.633 & 0.701 & 0.688 & 0.600 & 0.641 \\
	DV3+~\cite{chen2018encoder}    & 0.542 & 0.492 & 0.499 & 0.495 & 0.556 & 0.498 & 0.451 & 0.473 & 0.601 & 0.500 & 0.532 & 0.516 \\
	DV3+/R18~\cite{chen2018encoder} & 0.658 & 0.688 & 0.699 & 0.693 & \textbf{0.701} & 0.712 & 0.726 & 0.719 & 0.713 & 0.720 & 0.753 & 0.736 \\
	DRFN~\cite{wu2021document}     & \textbf{0.699} & \textbf{0.712} & \textbf{0.706} & \textbf{0.709} & 0.698 & \textbf{0.726} & \textbf{0.769} & \textbf{0.747} & \textbf{0.721} & \textbf{0.746} & \textbf{0.796} & \textbf{0.770} \\ \bottomrule
	\end{tabular}
	}
	\end{table*}

\subsection{Implementation Details}
{The DL-GDD framework is implemented using PyTorch. DLG was trained with a learning rate of $1e^{-5}$ using PubLayNet's label. To ensure layout diversity in the generated samples, we incorporated Magazine~\cite{zhengsig19} data to train the DLG, which resulted in 6000 document samples. Next, we used the DED to identify the elements in the generated layouts. Subsequently, we employed a trained document quality discriminator to evaluate the quality of the generated samples. From the evaluated samples, 3265 high-quality document images were selected. For style guidance, we trained a style discriminator using each target dataset (CS-150, DSSE-200, and CDSSE) and selected the images closest to these target datasets from 3265 high-quality pictures. The selected images are numbered 2135, 1620, and 1732 for CS-150, DSSE-200, and CDSSE, respectively. These images were used as inputs for the DL-GDD to generate additional samples until a total of 3000 generated samples were obtained. In summary, the training set consists of 335,703 samples in total, the fine-tuning dataset (after quality assessment) comprises 3265 samples, and the fine-tuning dataset for the cross-domain data consists of 3000 samples. We selected three representative datasets for evaluation and implemented certain classical methods to compare the effectiveness of the proposed methods.}

\subsection{{Datasets}}
\noindent
\textbf{{PubLayNet}.} The {PubLayNet}~\cite{zhong2019publaynet} is a large-scale DLA dataset. PubLayNet automatically annotates data from the PubMed center by matching XML representations. PubLayNet contains 360,000 annotated document images, and the document style pertains to scientific papers.

\vspace{0.08in}
\noindent
\textbf{DSSE-200.} The DSSE-200~\cite{yang2017learning} {is a dataset that focuses on complex samples. The DSSE-200 is composed of 200 pictures of manually annotated documents. It has a more decadent style and contains more document sources such as PPT pages, old newspapers and magazine pages, article pages, and scanned images.}

\vspace{0.08in}
\noindent
\textbf{CS-150.} The CS-150~\cite{clark2015looking} dataset focuses on academic documents. The data originates from 150 articles and were divided into 1175 pictures. CS-150 is grouped into three categories: figures, tables, and others.

\vspace{0.08in}
\noindent
\textbf{CDSSE.} CDSSE (Complex Document Semantic Structure Extraction)~\cite{wu2023drfn} is a large and complex non-Manhattan layout dataset collected and manually annotated by our team. The dataset comprised 500 document images. Compared to other DLA datasets, the CDSSE emphasizes the detailed annotation of the document layout. This includes the use of multiple rectangular boxes for more fine-grained division in several non-Manhattan layouts. Notably, the DSSE introduced the concept of open tables into this dataset.

\vspace{0.08in}
\noindent
\textbf{The gap between the datasets.}
{Several common dataset examples are shown in} Fig.~\ref{F_Dataset}, and labeled data are visualized. PubLayNet is an academic document dataset, and CS-150 is an academic paper dataset; however, there are specific style differences between the two data sets. CS-150 focuses on computer articles, whereas PubLayNet focuses on medical articles. Several other datasets are different from PubLayNet; therefore, from the perspective of datasets, it is necessary to consider the introduction of transitional data into the cross-domain process.

\subsection{Qualitative Results}

\noindent
\textbf{{PubLayNet} $ \to $ CS-150.} (Table~\ref{T_CS}).
We conducted three experiments to test the effect of cross-domain document layout analysis (DLA) on the CS-150 dataset. In the first experiment, we used document pages generated by PubLayNet and DED as the training set (33,5703 + 3265). In the second experiment, PubLayNet and document pages selected through quality screening were used as the training set (33,5703 + 3000). Finally, PubLayNet was used for training in the third experiment, with document pages guided by the cross-domain style as the training set (33,5703 + 3000). The results are summarized in Table~\ref{T_CS}.}
We observed that the quality screening strategy results were significantly better than those obtained directly from the generated samples.

To a certain extent, the results after using document-style migration are better than those of the previous two methods (because in CS-150, background and text are both considered. It is a non-picture area, and hence, the accuracy rate will ﬂuctuate). The F1 score improved by an average of 8\% compared with the samples generated using only DED. The best model was used to process the training samples, and the visualization results are shown in Fig.~\ref{F_RCS}.

\begin{figure}[t]
	\centering
	\includegraphics[width=\linewidth]{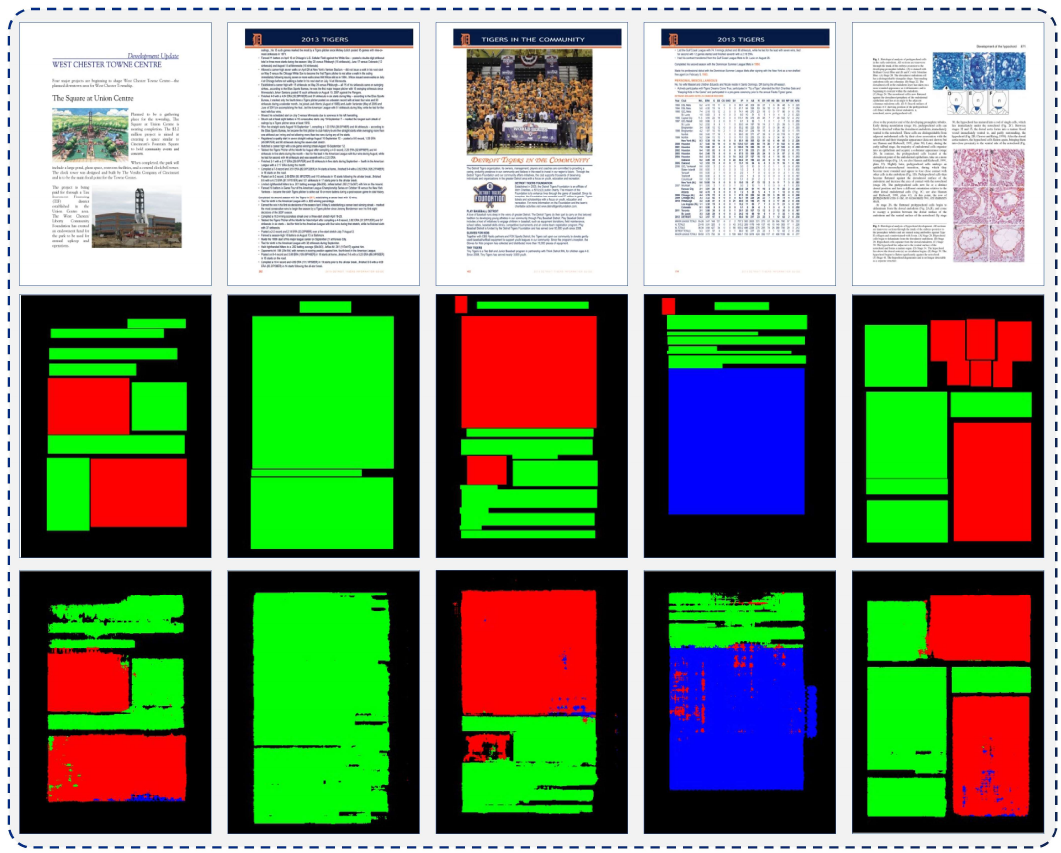}
	\caption{The DSSE-200 real documents and their corresponding segmentation. Top: original. Middle: ground-truth. Bottom: predictions. Segmentation label colors are: \colorbox{red}{figure}, \colorbox{blue}{\color{white}{table}}, \colorbox{green}{text} and \colorbox{black}{\color{white}{background}}.
	}
	\label{F_RDSSE}
\end{figure}

\vspace{0.08in}
\noindent
\textbf{{PubLayNet} $ \to $ DSSE-200.} (Table~\ref{T_DSSE}).
Following the CS-150, we conducted experiments on DSSE and used the entire DSSE as the test set without fine-tuning. Table~\ref{T_DSSE} presents the results of the study. The results showed that those obtained using the document style cross-domain migration are significantly better than the other two methods. The experimental results clearly indicated that document-style cross-domain migration leads to a substantial improvement in F1 score, with an average increase of 16.6\% compared with samples generated solely through DED. The visualization results are presented in Fig.~\ref{F_RDSSE}.

\begin{figure}[t]
	\centering
	\includegraphics[width=\linewidth]{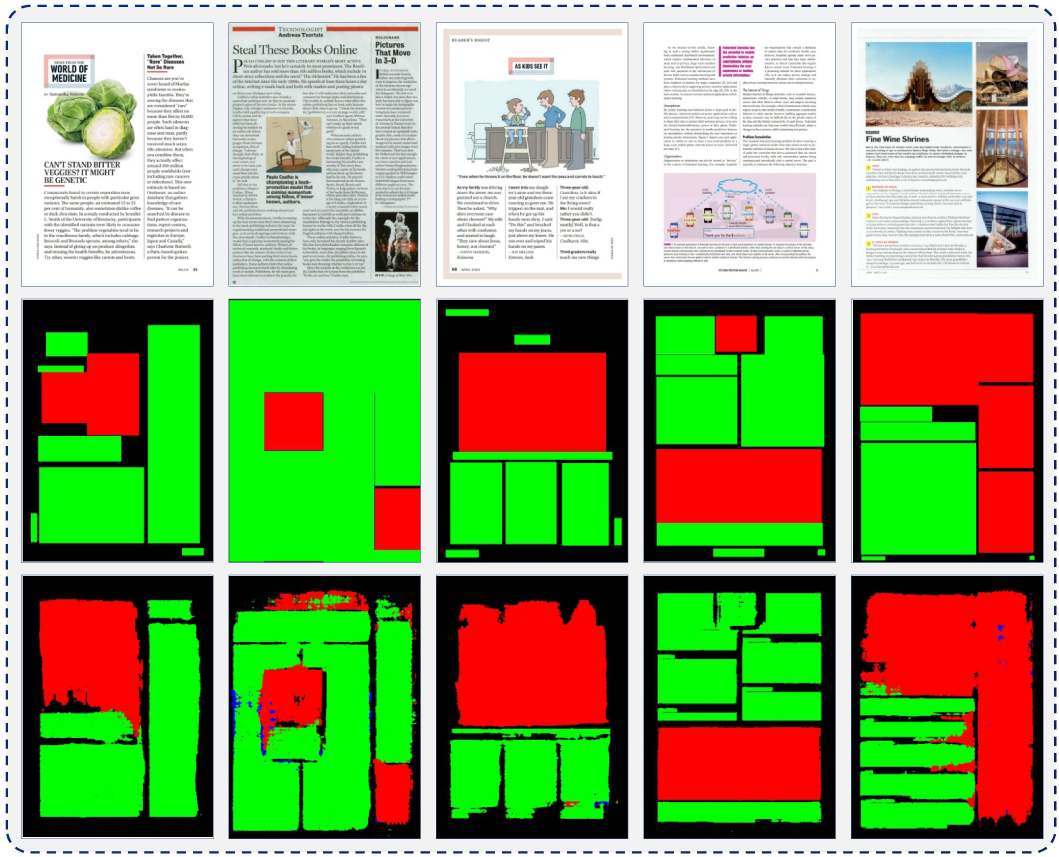}
	\caption{ The CDSSE real documents and their corresponding segmentation. Top: original. Middle: ground-truth. Bottom: predictions.
Segmentation label colors are: \colorbox{red}{figure}, \colorbox{blue}{\color{white}{table}}, \colorbox{green}{text} and \colorbox{black}{\color{white}{background}}.}
	\label{F_RCDSSE}
\end{figure}

\vspace{0.08in}
\noindent
\textbf{{PubLayNet} $ \to $ CDSSE.} (Table~\ref{T_CDSSE}).
Following the CS-150, we conducted experiments on CDSSE and used all CDSSE as the test set without fine-tuning. Table~\ref{T_CDSSE} presents the results of the study. The results obtained by using document style cross-domain migration were significantly better than those of the other two methods. The results presented in Table~\ref{T_CDSSE} demonstrate that, upon employing document-style cross-domain migration, the F1 score almost surpassed 60\%, signifying an average enhancement of 11.9\% compared with the samples generated solely using DED. Because the CDSSE is a more intricate dataset and introducing open tables is a new factor, the recognition effectiveness remains limited. The visualization results are shown in Fig.~\ref{F_RCDSSE}.

\begin{figure*}[t]
	\centering
	\includegraphics[width=\linewidth]{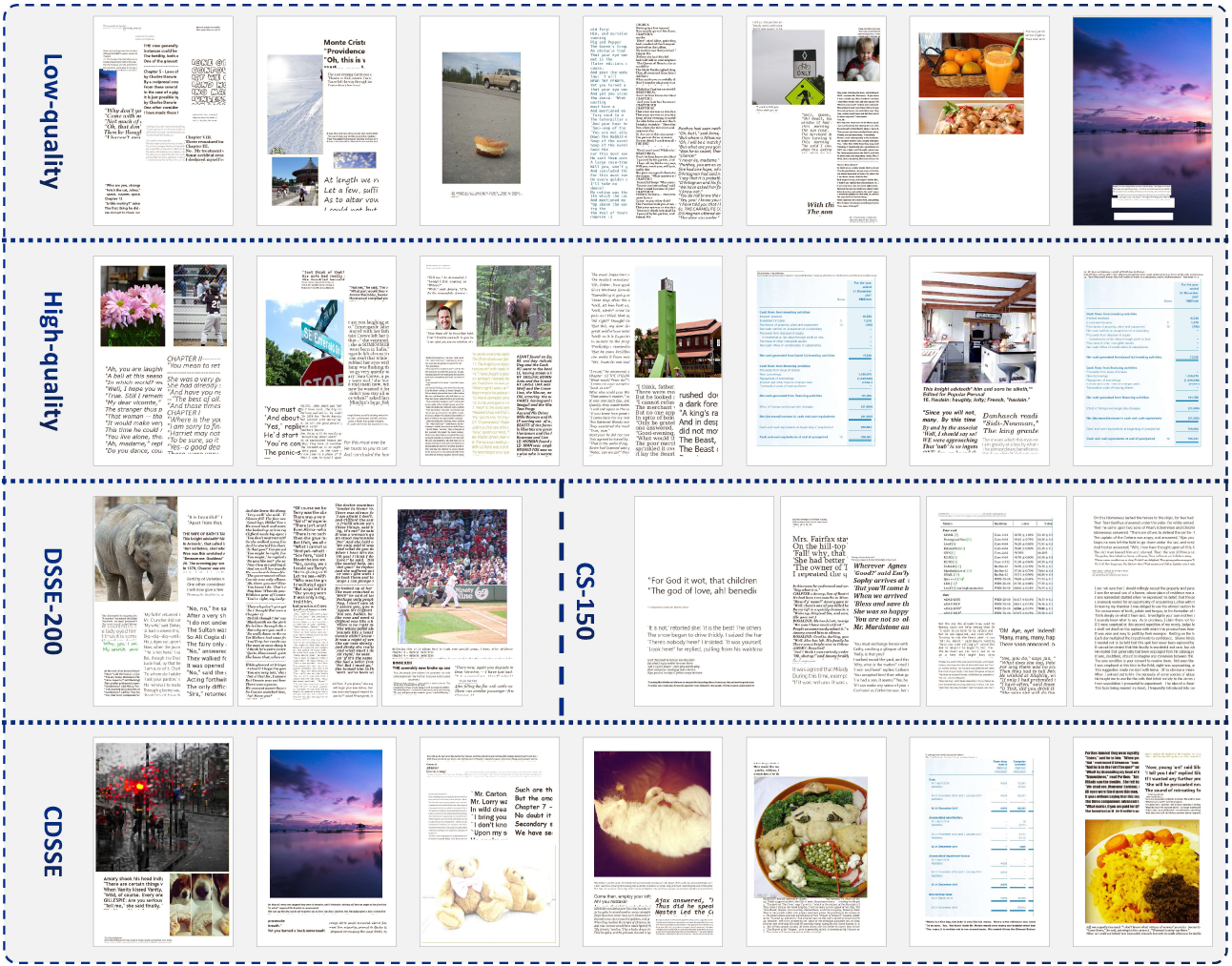}
	\caption{The image processed by DL-GDD, the first and {second rows include the} images filtered by the document quality discriminator, and the third and {fourth rows have the} images guided by the cross-domain style guide.}
	\label{F_Ab}
\end{figure*}

\subsection{Ablation Study}
\label{sec:Ablation}

We conducted a series of comparative experiments using a benchmark dataset to verify the effectiveness of the proposed approach.

\vspace{0.08in}
\noindent
\textbf{Document Layout Generator \& Document Elements Decorator}
In Fig.~\ref{F_Ab}, we present several examples of document layouts generated using DLG and DED. These examples demonstrate the adequate performance of the DLG and DED components. Furthermore, we implemented a fault-tolerant mechanism to endow our components with robustness, enabling them to handle unexpected situations that may arise (across 6000 document layouts). To optimize the processing time, we devised a multithreaded processing approach.

\vspace{0.08in}
\noindent
\textbf{Document Style Discriminator for Quality Assessment.}
In Fig.\ref{F_Ab}, the first and second rows display samples classified as low-quality and high-quality by DSD. Upon examining the images, it is evident that the high-quality samples exhibit more coherence and rational layout arrangements than their low-quality counterparts. Moreover, the results observed in Tables \ref{T_CS}, ~\ref{T_DSSE}, and ~\ref{T_CDSSE} further validate these findings. Despite using a smaller dataset for quality assessment compared to the directly generated samples, we noted that employing quality assessment models yields significantly improved results. This underscores the significant impact of the DSD module that was employed for quality assessment.

\vspace{0.08in}
\noindent
\textbf{Document Style Discriminator for Cross-domain Document Style Guidance.}
In Fig.~\ref{F_Ab}, we illustrate examples of the implementation of document cross-domain style guidance in the third and fourth rows. Notably, the samples employing CS-150 for cross-domain guidance predominantly exhibited an academic style and DSSE-200 for the cross-domain guidance results in the widest variety of styles, whereas the samples using CDSSE primarily adopted a magazine layout. Furthermore, the samples from the open table category were selected from the dataset. Upon evaluating the results presented in Tables ~\ref{T_CS}, ~\ref{T_DSSE}, and ~\ref{T_CDSSE}, it was apparent that the outcomes achieved through document-style cross-domain guidance surpassed those obtained using the other two methods in terms of quality and effectiveness.

\section{Conclusions}
\label{sec:conclusion}
In this study, we propose an unsupervised document-style cross-domain guidance framework called DL-GDD, which is composed of three components. 
The DL-GDD integrates document-page quality assessment and document cross-domain analysis for the first time. Specifically, we propose an unsupervised method that does not rely on annotated data to narrow the gap in the cross-domain DLA. 
We also extended the proposed method to make it suitable for the quality assessment of the DLA. We proposed a document element decorator to connect the DLG and document layout quality assessment components. 
This component implements document-element decoration under aesthetic specifications. Extensive experiments on document layout analysis benchmarks demonstrated the superior performance of the proposed method.

\bibliographystyle{IEEEtran}
\bibliography{egbib}

\end{document}